\documentclass{article}

\usepackage[final]{corl_2026} %

\usepackage{amssymb}
\usepackage{amsmath}
\usepackage{graphicx}
\usepackage{xcolor}
 \usepackage{booktabs}
 \usepackage{tabularx}
\usepackage{multirow}
\usepackage{wrapfig}

\usepackage{tcolorbox}
\tcbuselibrary{breakable, skins}
\newtcolorbox{promptbox}[1][]{colback=gray!8, colframe=gray!30,
  fonttitle=\bfseries, title=Prompt, left=4pt, right=4pt, breakable,
  coltitle=black, #1}

\usepackage{hyperref}
\renewcommand*{\doi}[1]{\textsc{doi}: \href{https://doi.org/#1}{#1}}

  \hypersetup{pdftitle={Asking Smarter Questions: Targeted Active Preference Learning for Underspecified Rewards},
              pdfauthor={Helena Merker, Nick Walker, Andreea Bobu}}

\usepackage{titlesec}

\titlespacing*{\paragraph}
{0pt}                   %
{0.0ex}       %
{1em}                    %

\title{Asking Smarter Questions: 
Targeted Active Preference Learning for Underspecified Rewards}

\author{
  Helena Merker, Nick Walker, Andreea Bobu\\
  Massachusetts Institute of Technology \\
  \{hmerker, nswalker, abobu\}@mit.edu
}

\begin{document}
\maketitle

\begin{abstract} 
    Reward learning typically assumes the human's input covers everything that matters for the task. In practice, humans can only attend to so much at once, so their teaching naturally captures some features---or aspects of the task---better than others. Prior work has shown how the robot can identify which features were poorly supervised directly from its teaching data. But the current solution is to ask the human to try again, putting the burden on someone who already struggled to convey those features. We propose shifting this burden to the robot.
    Our key insight is that the same information that tells the robot what it doesn't know can help it ask better questions. Preference queries offer a low-effort channel for resolving this uncertainty, but only if the robot knows where to direct them and the human knows what to attend to when answering. We act on this insight in two ways: we build an informed prior that targets preference queries at the features that need resolving, and we accompany each query with an explanation that focuses the human's attention on exactly those features.
    In simulation, the informed prior beats the standard active preference learning baseline by up to $30.8\%$ in the low-query regime, where reducing human effort matters most. In a user study, the informed prior and explanations each significantly improve reward recovery and amplify each other.
\end{abstract}


\section{Introduction}

When a human teaches a robot how to do a task, not every relevant aspect of the task---or \textit{feature}---gets equal attention. For instance, while teaching a robot how to carry a knife across your cluttered kitchen, you might do a good job avoiding the hot stovetop or the rack hanging overhead, but struggle to consistently keep the knife away from every person or obstacle. Because humans have limited working memory and can't focus on everything at once~\citep{sweller1988cognitive, baddeley1992working}, some features get carefully attended to while others are left \textit{underspecified}. Recent work shows how the robot can detect, directly from its teaching data, which features were poorly supervised or entirely irrelevant to the task~\cite{merker2026robots,peng2023DFA}. We go a step further and ask: how should the robot work with the human to \textit{resolve} that underspecification?

One approach is to ask the human for more demonstrations focused on the underspecified features~\cite{merker2026robots}. But this places the burden on the human, who already struggled with attending to those features. We propose to shift this burden onto the robot: rather than ask the human to try again, the robot should itself ask targeted questions that resolve what it is uncertain about.

Preference queries, where the robot generates behavior options and the human just picks their preferred one, offer this kind of low-effort channel~\citep{sadigh2017active, biyik2019asking, peng2024pragmaticfeaturepreferences}. %
Active learning approaches select queries to maximally reduce uncertainty under a distribution over the human's reward~\cite{biyik2019asking}. But when that distribution is initialized from imperfect teaching data~\citep{biyik2022learningreward}, it inherits not just what the feedback got right but also its errors on the underspecified features. The robot then spends its early queries correcting a misaligned distribution before it can make progress on what it actually needs to learn.

\begin{figure}[t]
    \centering
    \includegraphics[width=\linewidth]{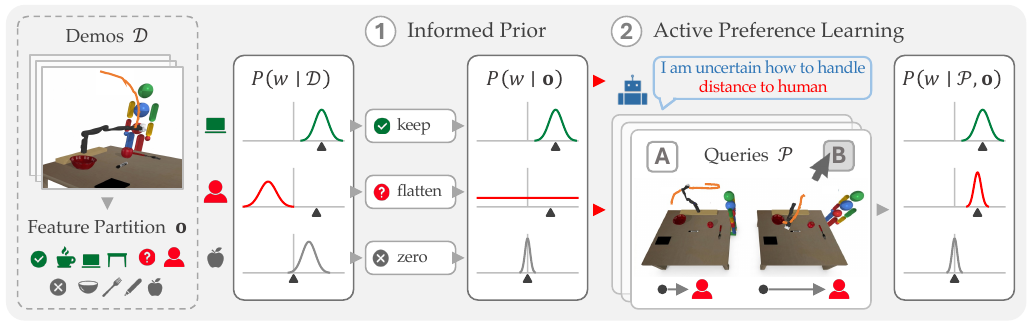}
    \caption{
    A user teaches a robot to carry a mug of coffee but pays little attention to how close it gets to the human. Using prior work~\cite{merker2026robots}, the robot diagnoses from the demonstrations $\mathcal{D}$ which features were well-supervised (green), underspecified (red), or irrelevant (gray), and uses this partition $\mathbf{o}$ in two ways. (1) The robot shapes its belief $P(w \mid \mathcal{D})$ (true weights marked by $\blacktriangle$) into an informed prior $P(w \mid \mathbf{o})$: the robot stays confident about the laptop, doesn't commit to anything about the human, and assumes the apple doesn't matter. (2) The robot tells the user what to focus on. Since the robot is only unsure about distance to human, it asks about two trajectories that differ in exactly that, and each answer sharpens $P(w \mid \mathcal{P}, \mathbf{o})$ toward what it still needs to learn.
    }
    \label{fig:main_figure}
\end{figure}

Our key insight is that \textit{the same information that tells the robot what it doesn't know can help it ask better questions}. The robot's diagnosis of what past teaching missed serves two roles. %
First, we use it to build an informed prior over the human's reward: a Gaussian that is confident on the well-supervised features, flat on the underspecified ones, and near zero on irrelevant ones. This ensures preference queries are automatically directed at what needs resolving.
Second, we use it to supplement each query with an explanation that tells the human which features to focus on when answering, making the robot's learning process transparent~\citep{habibian2022heres}. 
In \autoref{fig:main_figure}, the robot communicates uncertainty about distance to human, and its query presents two options that vary along that feature. Together, these two components align what the robot asks with what the human attends to.

Our contributions are: 
\textbf{(1)} an informed prior for active preference learning that corrects what the teaching data got wrong, automatically targeting queries at the underspecified features; 
\textbf{(2)} an explanation mechanism that directs the human's attention to those same features when answering, aligning what the robot asks with what the human attends to; and 
\textbf{(3)} empirical validation in two simulated 7-DoF manipulation domains and a user study, where both components independently improve reward recovery and amplify each other, with the largest gains in the low-query regime.

\vspace{-0.5mm}
\section{Related Work}

\paragraph{Learning from Imperfect Teaching}
Reward learning from human feedback typically assumes the teacher is optimal or noisily optimal~\citep{abbeel2004apprenticeship, ziebart2008maximum}, but human input is routinely suboptimal. Some methods build richer models of the human, capturing systematic or structured deviations from optimality~\citep{laidlaw2022boltzmann, shah2019feasibility}. Others treat suboptimality as localized failures of supervision rather than uniform noise: the robot reasons about how well its hypothesis space explains the human's input and flags what it cannot account for~\citep{bobu2018misspecified, bobu2020quantifying,zurek2021confidence}, or detects from the teaching data which features were well-supervised versus underspecified~\citep{merker2026robots, peng2023DFA,lourenco2023diagnosing}. 
Rather than resolving the diagnosed gap by requesting more data, we use the diagnosis to drive active preference queries, shifting resolution onto the robot. 

\paragraph{Active Preference Elicitation}
Preference queries, which ask the human to choose between trajectories, are a low-effort channel for reward learning. The robot can select each query to maximize information gain, minimizing the comparisons the human must answer~\citep{biyik2019asking}. Prior work extends this loop in various ways: warm-starting the reward belief from demonstrations~\citep{biyik2022learningreward}, enriching queries to ask which features make a trajectory preferable~\citep{peng2024pragmaticfeaturepreferences}, and using language models to select queries and ease answering~\citep{mahmud2025maple}. 
From~\citet{biyik2022learningreward} we inherit the initialization of the reward belief from demonstrations. But rather than treating all features alike, we shape that belief to encode where teaching was reliable, letting the information-gain criterion target the underspecified features.

\section{Problem Setup}

We address the problem of learning a reward function from human input when the teaching signal leaves some features underspecified. %
Recent work shows that robots can detect underspecified features from statistical patterns in the teaching data~\cite{merker2026robots}. We use this detection and ask: how should the robot resolve underspecified features? 
We propose using that knowledge in two ways: to correct the imperfect posterior into an informed prior, and to direct human attention during preference queries.

\subsection{Preliminaries}

The robot's goal is to learn a reward function that captures how the human wants the task performed. We adopt a linear reward parameterization~\citep{abbeel2004apprenticeship,ziebart2008maximum} over state features. Let $\phi: \mathcal{S} \rightarrow \mathbb{R}^k$ be a feature mapping that extracts a $k$-dimensional vector from each state $s \in \mathcal{S}$, and let $w \in \mathbb{R}^k$ be the reward weight vector with $\|w\|_2 \leq 1$. The per-state reward is $R_w(s) = w^\top \phi(s)$, and the cumulative reward of a trajectory $\xi = (s^0, s^1, \ldots, s^T) \in \Xi$ aggregates state-level contributions
$R_w(\xi) = \sum_{s^t \in \xi} w^\top \phi(s^t) = w^\top \phi(\xi)$,
where $\phi(\xi) = \sum_{s^t \in \xi} \phi(s^t)$ accumulates features along the trajectory. We write $w^\star$ for the human's true preference weights, which are unknown to the robot. The features $\phi$ may be hand-crafted~\citep{ziebart2008maximum}, learned~\citep{bobu2022inducing}, or extracted from a large language model given a high-level task description~\citep{peng24algae}; we assume access to a candidate feature set.

\subsection{An Imperfect Posterior}
\label{sec:imperfect_posterior}

We assume the robot has access to a dataset $\mathcal{D}$ of human teaching data, and maintains a posterior $P(w \mid \mathcal{D})\propto P(\mathcal{D} \mid w)\, P(w)$ over the reward weights. 
In general, the likelihood $P(\mathcal{D} \mid w)$ can be instantiated for any teaching modality $\mathcal{D}$, including demonstrations~\citep{abbeel2004apprenticeship,ziebart2008maximum}, preferences~\citep{sadigh2017active,biyik2022learningreward}, corrections~\citep{bajcsy2017learning}, or natural language~\citep{sumers2021learning}. 
For exposition, we follow \textit{DemPref}~\citep{biyik2022learningreward} and initialize the reward belief from Boltzmann-rational demonstrations~\citep{jaynes1957infotheory,von1945theory,baker2007goal}, which we then refine with preference queries. The resulting posterior $P(w \mid \mathcal{D})$ is rarely uniform in quality across the components of $w$. Teaching a task that requires attending to multiple objectives at once is cognitively demanding; humans inevitably focus on some features at the expense of others. The demonstrations are thus not driven by $w^\star$ equally across all features, leaving the posterior misleading along the underspecified dimensions. Prior work~\cite{merker2026robots,peng2023DFA} showed how the robot can detect, from $\mathcal{D}$, which features were well-optimized and which were underspecified. The candidate feature set may also contain task-irrelevant features, whose posterior reflects spurious correlations rather than the human's intent.

\paragraph{Feature Partition} From~\citet{merker2026robots} we inherit the variance-based detection of underspecified features and the LLM-based filtering of task-irrelevant ones, which together partition the $k$ features into three disjoint subsets, collectively denoted by the partition $\mathbf{o}$: 1) $\phi^o$: the task-relevant features that are well-optimized in $\mathcal{D}$, 2) $\phi^u$: the task-relevant features that are underspecified, and 3) $\phi^\varepsilon$: the features 
irrelevant to the task. We write $w^o$, $w^u$, $w^\varepsilon$ for the blocks of $w$ corresponding to $\phi^o$, $\phi^u$, $\phi^\varepsilon$. 

\paragraph{Refining the Prior with Active Feedback} 
The partition $\mathbf{o}$ summarizes what $\mathcal{D}$ reveals: $P(w \mid \mathcal{D})$ is trustworthy on $\phi^o$, misleading on $\phi^u$, and uninformative on $\phi^\varepsilon$. Our goal is to elicit additional human feedback that sharpens our estimate of $w^\star$ along the underspecified dimensions. 
One approach is to passively request additional demonstrations of the features in $\phi^u$, returning the burden of producing behavior that exercises those features to the human~\cite{merker2026robots}. Instead, we adopt an active preference learning framework~\cite{biyik2019asking}, where the robot itself generates queries to resolve its uncertainty. 

Standard active preference learning~\cite{biyik2019asking} uses $P(w \mid \mathcal{D})$ as the prior; however, its deficiencies in $\phi^u$ and $\phi^\varepsilon$ force early queries to first correct the prior before they can resolve the underspecified features. We use $\mathbf{o}$ in two ways to address this. First, we construct a corrected distribution $P(w \mid \mathbf{o})$ by modifying $P(w \mid \mathcal{D})$ block-wise: retaining what $\mathcal{D}$ revealed about $\phi^o$, flattening the prior on $\phi^u$, and concentrating it near zero on $\phi^\varepsilon$ (see \autoref{sec:prior}).
We use $P(w \mid \mathbf{o})$ as the \textit{prior} for active preference learning, directing the robot's queries at its uncertainty along $\phi^u$ (\autoref{sec:active_pref}).
Second, we adapt the explanation of~\citet{merker2026robots}, reframed to instruct the human to attend to $\phi^u$ when answering queries, so that their responses inform the specific features that need resolving (\autoref{sec:likelihood}).

\section{Method}

Our approach reuses the robot's diagnosis of which features past teaching left underspecified to elicit targeted preference feedback. First, we construct an informed prior $P(w \mid \mathbf{o})$ that corrects $P(w \mid \mathcal{D})$ block-wise, concentrating uncertainty on $\phi^u$. Second, we pair each query with an explanation that focuses the human's attention on $\phi^u$ when answering, so their responses inform those features.

\subsection{The Informed Prior}
\label{sec:prior}

We now describe how to construct $P(w \mid \mathbf{o})$ in place of the standard initialization from $P(w \mid \mathcal{D})$.
We want a corrected version of $P(w \mid \mathcal{D})$ that leaves $\phi^o$ untouched, is uninformative on $\phi^u$, and concentrates near zero on $\phi^\varepsilon$. This requires a parametric form for $P(w \mid \mathcal{D})$ amenable to block-wise edits. We represent $P(w \mid \mathcal{D})$ as a Gaussian, whose parameters we modify to obtain $P(w \mid \mathbf{o})$.

\paragraph{Gaussian Approximation} %
To approximate $P(w \mid \mathcal{D})$, we draw samples via Metropolis-Hastings with the unnormalized posterior $P(\mathcal{D} \mid w)\, P(w)$~\citep{biyik2019asking}, %
and fit a Gaussian $\mathcal{N}(\mu, \Sigma)$ to them. 
This yields the best Gaussian summary of $P(w \mid \mathcal{D})$, i.e. the closest in KL divergence~\citep{minka2001expectation}.
However, $\mathcal{N}(\mu, \Sigma)$ inherits the deficiencies of $P(w \mid \mathcal{D})$: only the blocks $\mu^o$ and $\Sigma^{oo}$ for the well-optimized features $\phi^o$ are informative, while $\mu^u$ and $\Sigma^{uu}$ came from demonstrations where $\phi^u$ was underspecified, and $\mu^\varepsilon$ and $\Sigma^{\varepsilon\varepsilon}$ came from spurious correlations. The block-wise modification below corrects these.

\paragraph{Constructing the Informed Prior}
The informed prior $P(w \mid \mathbf{o})$ takes a different form on each of the three blocks of $\mathbf{o}$. On $\phi^o$, $P(w \mid \mathbf{o})$ should preserve the information about $w^{\star, o}$ already contained in $P(w \mid \mathcal{D})$, since the demonstrations are informative on these features. On $\phi^u$, $P(w \mid \mathbf{o})$ should be uninformative about $w^{\star, u}$ since the human did not attend to those features; this way, the preference queries in \autoref{sec:active_pref} are responsible for setting $w^u$. On $\phi^\varepsilon$, $P(w \mid \mathbf{o})$ should concentrate $w^\varepsilon$ near zero. %

We construct $P(w \mid \mathbf{o})$ as a Gaussian density on $w$:
\begin{equation}
    P(w \mid \mathbf{o}) \;\propto\; \exp\!\Bigl(-\tfrac{1}{2}(w - \tilde\mu)^\top \tilde\Lambda\, (w - \tilde\mu)\Bigr),
    \label{eq:informed_prior}
\end{equation}
with mean $\tilde\mu$ and precision matrix $\tilde\Lambda$ 
obtained by per-block modification of $\mu$ and $\Sigma$. %
The mean is $\tilde\mu^o = \mu^o, \ \tilde\mu^u = 0, \ \tilde\mu^\varepsilon = 0$,
and the precision is block-diagonal across $\phi^o, \phi^u, \phi^\varepsilon$:
$
    \tilde\Lambda^{oo} = (\Sigma^{oo})^{-1}, \
    \tilde\Lambda^{uu} = 0, \
    \tilde\Lambda^{\varepsilon\varepsilon} = \mathrm{diag}\bigl((\lambda_l)_{l \in \phi^\varepsilon}\bigr).
$
Setting $\tilde\mu^o = \mu^o$ and $\tilde\Lambda^{oo} = (\Sigma^{oo})^{-1}$ carries over the demonstration-derived information about $w^{\star, o}$ from $P(w \mid \mathcal{D})$. %
Setting $\tilde\Lambda^{uu} = 0$ makes $P(w \mid \mathbf{o})$ flat in $w^u$, so all information about $w^u$ comes from the preference queries. 
On $\phi^\varepsilon$, we set $\tilde\mu^\varepsilon = 0$ and $\tilde\Lambda^{\varepsilon\varepsilon} = \mathrm{diag}((\lambda_l)_{l \in \phi^\varepsilon})$, where $\lambda_l \geq 0$ sets the concentration strength for $\phi_l$ proportionally to how likely that feature is to be irrelevant. This concentrates $w^\varepsilon$ near zero, encoding the belief that $w^{\star,\varepsilon} \approx 0$. In practice, we use an LLM queried with the task description to estimate per-feature irrelevance, and describe the full computation of $\lambda_l$ in \autoref{app:experimental-details}.

\subsection{Active Preference Learning}
\label{sec:active_pref}

Pairwise preference queries are a low-effort channel for reward inference~\citep{sadigh2017active}: the robot presents two trajectories $Q_i = \{\xi^A_i, \xi^B_i\}$ and asks the human which they prefer, $q_i \in Q_i$. The probability follows a Bradley-Terry model~\citep{bradley1952rank}, which typically assumes the human responds purely based on which trajectory has higher reward, with no explicit model of attention. Inspired by the Boltzmann rationality model used for demonstrations~\citep{jaynes1957infotheory,merker2026robots}, we introduce a per-query, per-feature rationality vector $\beta_i \in \mathbb{R}^k_{\geq 0}$ that controls how much each feature influences the human's response. Each component $\beta_{i,l}$ modulates the contribution of feature $\phi_l$ on query $i$: setting $\beta_{i,l} = 0$ removes $\phi_l$ from the likelihood entirely, encoding that the human did not attend to that feature when answering, so the choice is carried out as if that feature did not exist. This per-query indexing lets the robot direct what the human attends to on each query. We describe how $\mathbf{o}$ sets the values of $\beta_{i}$ in \autoref{sec:likelihood}. 
The preference probability is then: 
\begin{equation}
    P(q_i = \xi^j_i \mid Q_i, w, \beta_i) = \frac{\exp\bigl(\sum_{l=1}^{k} \beta_{i,l}\, w_l\, \phi_l(\xi^j_i)\bigr)}{\exp\bigl(\sum_{l=1}^{k} \beta_{i,l}\, w_l\, \phi_l(\xi^A_i)\bigr) + \exp\bigl(\sum_{l=1}^{k} \beta_{i,l}\, w_l\, \phi_l(\xi^B_i)\bigr)}, \quad j \in \{A, B\}\enspace.
    \label{eq:bt_vector}
\end{equation}

The robot sets $\beta_i$ based on the partition $\mathbf{o}$ before issuing each query, encoding an assumption about which features the human will attend to when answering (see \autoref{sec:likelihood}). 
The data collected from $K$ such queries is $\mathcal{P} = \{(Q_i, q_i)\}_{i=1}^{K}$, and since $\beta_i$ is determined by $\mathbf{o}$, the full posterior is:
\begin{equation}
    P(w \mid \mathcal{P}, \mathbf{o}) \;\propto\; P(\mathcal{P} \mid w, \mathbf{o}) P(w \mid \mathbf{o}) = \Biggl[\prod_{i=1}^{K} P(q_i \mid Q_i, w, \beta_i)\Biggr] \, P(w \mid \mathbf{o})\enspace.
    \label{eq:informed_posterior}
\end{equation}

Given this posterior, the robot must decide which query to ask next. At each step, it selects the query that maximally reduces uncertainty about $w$ under the current posterior, i.e. the query that maximizes the mutual information between the human's response and the reward weights~\citep{biyik2019asking}:
\begin{equation}
    Q_i^{*} = \arg\max_{Q_i} I\bigl(w; q_i \mid Q_i, \beta_i\bigr) = \arg\max_{Q_i} H\bigl(w \mid Q_i, \beta_i\bigr) - \mathbb{E}_{q_i}\bigl[H\bigl(w \mid q_i, Q_i, \beta_i\bigr)\bigr],
    \label{eq:ig_abstract}
\end{equation}
where $I$ is the mutual information and $H$ denotes Shannon entropy~\citep{cover2012elements}. We approximate the posterior with $M$ samples $\Omega = \{w_1, \ldots, w_M\}$. Letting $\doteq$ denote asymptotic equality as $M \to \infty$~\citep{biyik2019asking}, the criterion becomes
\begin{equation}
    Q_i^{*} \doteq \arg\max_{Q_i} \frac{1}{M} \sum_{q_i \in Q_i} \sum_{w \in \Omega} P(q_i \mid Q_i, w, \beta_i) \log_2 \frac{M \cdot P(q_i \mid Q_i, w, \beta_i)}{\sum_{w' \in \Omega} P(q_i \mid Q_i, w', \beta_i)}.
    \label{eq:ig_sample}
\end{equation}
Because the informed prior $P(w \mid \mathbf{o})$ is flat on $w^u$ and tightly constrained on $w^o$ and $w^\varepsilon$, uncertainty concentrates along $w^u$. %
The information gain criterion therefore automatically directs queries toward trajectories that differ along $\phi^u$, which are the components we intend for the preferences to inform. 

\subsection{Directing Attention with Explanations}
\label{sec:likelihood}

The second use of $\mathbf{o}$ is directing the person's attention during preference queries. The partition $\mathbf{o}$ is a record of what past teaching revealed, i.e. which features were well-optimized, underspecified, or irrelevant. We now repurpose it as prospective guidance, telling the human which features to attend to when answering queries. In that sense, $\mathbf{o}$ determines $\beta_i$, the rationality vector that governs how the human's response enters the likelihood.
Without an explanation, the human is assumed to attend to every feature equally, corresponding to
$\beta_i = \beta_0 \mathbf{1}$, with the same positive value $\beta_0$ on every component. Alternatively, the robot can provide an explanation that directs the human's attention to the underspecified features $\phi^u$ (e.g., ``I'm uncertain about how to handle [features in $\phi^u$]. Please focus on these features when comparing the two trajectories.''). This is how $\mathbf{o}$ determines $\beta_i$: features in $\phi^u$ receive rationality $\beta_0$ while $\phi^o$ and $\phi^\varepsilon$ are zeroed out, giving $\beta_i^u = \beta_0 \mathbf{1}$ and $\beta_i^o = \beta_i^\varepsilon = 0$. The Bradley-Terry probability in \autoref{eq:bt_vector}---and hence the likelihood $P(\mathcal{P} \mid w, \mathbf{o})$ in \autoref{eq:informed_posterior}---then depends on $w$ only through $w^u$. The trajectories therefore enter the query only through their $\phi^u$ features, so a query whose two trajectories agree on those features produces no information at all.

\section{Simulated Experiments}
\label{sec:simulated_experiments}

We evaluate our method in simulation, comparing the informed prior against four alternative priors, a targeted query objective, random querying, and three attention-directing explanation variants.

\subsection{Experimental Setup}

\begin{figure}
    \centering
    \includegraphics[width=0.85\linewidth]{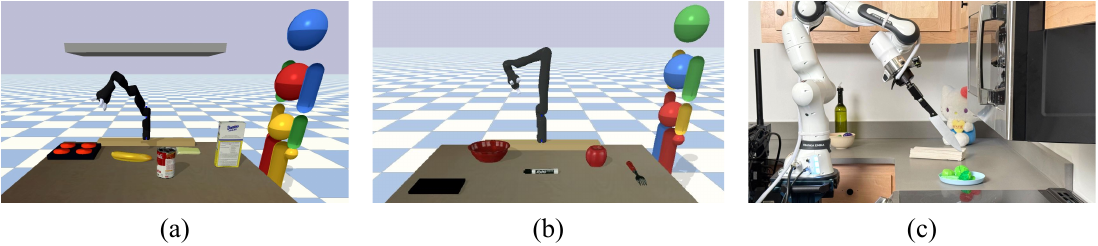}
    \caption{(a) \textit{Stove}: the robot carries a utensil near a stove and a human. (b) \textit{Coffee}: the robot carries a coffee cup near a laptop and a human. (c) A Franka Emika FR3 arm transports a knife in a kitchen.}
    \label{fig:environments}
\end{figure}

We use two scenes wherein a 7-DoF Jaco arm carries an object above a cluttered surface (\autoref{fig:environments}a--b). In \textbf{Stove}, the robot carries a utensil across a workspace with a stove, a human, an overhead rack, and four distractors (sponge, banana, soup can, sugar box). The task-relevant features are: (1) \textit{stove}, the $xy$-distance from the end-effector to the stove; (2) \textit{human}, the $xy$-distance to the human; (3) \textit{point}, the dot product between the end-effector's pointing direction and the unit vector toward the human; and (4) \textit{rack}, the $z$-distance below the rack. In \textbf{Coffee}, the robot carries a coffee cup across a workspace containing a laptop, a human, and four distractors (apple, bowl, fork, marker). The task-relevant features are: (1) \textit{laptop}, the $xy$-distance to the laptop; (2) \textit{table}, the $z$-distance of the end-effector above the table; (3) \textit{human}, the $xy$-distance to the human; and (4) \textit{coffee}, the projection of the end-effector's vertical axis onto the world up-axis, measuring how upright the cup remains. For each environment, we construct 50,000 trajectories from 5,000 random start-goal end-effector pairs, linearly interpolated in joint space and perturbed by 10 random smooth deformations each. Each trajectory has 21 waypoints encoding joint configurations and the $xyz$ positions and rotations of all joints. We split the bank 60-20-20\% into training, validation, and test sets. The simulated human draws its 10 initial demonstrations and all preference-query pairs from the training set.

We model the demonstrator as Boltzmann-rational with ground-truth weights $w^\star$ and a per-feature rationality vector $\beta \in \mathbb{R}^k_{\geq 0}$. A high $\beta_l$ produces demonstrations that reliably optimize $\phi_l$, while $\beta_l = 0$ simulates underspecification. Distractors are fixed to $w^{\star}_l = 0$. When answering queries without an explanation, the human uses $\beta_i = \beta_0 \mathbf{1}$, so responses follow the full $w^{\star}$. When the robot provides an explanation directing attention to $\phi^{\mathrm{exp}} \subseteq \phi$, the human's per-query rationality vector has nonzero entries only on the indices of $\phi^{\mathrm{exp}}$, so the response depends only on those features.

\paragraph{Methods}
(1)~\textit{Informed Prior} constructs $P(w \mid \mathbf{o})$ by block-wise modification of a Gaussian approximation of $P(w \mid \mathcal{D})$. The active query loop targets $P(w \mid \mathcal{P}, \mathbf{o})$ and selects queries by information gain.
(2)~\textit{Informed Prior + Informed Explanation} further sets $\phi^{\mathrm{exp}}$ to the predicted $\phi^u$, so the human attends to the same features the prior leaves unconstrained.
(3)~\textit{Oracle Prior} uses the true set of underspecified features.
(4)~\textit{Oracle Prior + Oracle Explanation} further sets $\phi^{\mathrm{exp}}$ to the true underspecified features.
(5)~\textit{Extra-Only Prior} is a version of the informed prior that does not predict underspecified features. It combines the predicted irrelevant features $\phi^\varepsilon$, represented as a diagonal Gaussian, with the demonstration likelihood. 
(6)~\textit{Standard} models posterior $P(w \mid \mathcal{P}, \mathcal{D})$ by combining the demonstration likelihood with a uniform prior over the unit ball and selects queries by information gain.
(7)~\textit{Standard + Oracle Explanation} sets $\phi^{\mathrm{exp}}$ to the true underspecified features, isolating the contribution of the explanation alone.
(8)~\textit{Targeted} uses the unmodified $P(w \mid \mathcal{D})$ and no explanation ($\beta_i = \beta_0 \mathbf{1}$), but selects queries by maximizing $I(w^u; q_i \mid Q_i, \beta_i)$ over the predicted $\phi^u$.
(9)~\textit{Gaussian Approximation} is Standard but with the Gaussian approximation of $P(w \mid \mathcal{D})$, indicating the penalty of summarizing the original posterior.
(10)~\textit{Random} draws queries uniformly at random. We report \textbf{normalized reward}: the average ground-truth reward of the top-$k$ test trajectories under the posterior mean, normalized by the range of ground-truth rewards across the test set. All experiments are repeated across $5$ seeds, use $k=20$, and we report mean and standard error.

\subsection{Results}

\begin{figure*}[t]
    \centering
    \includegraphics[width=\linewidth]{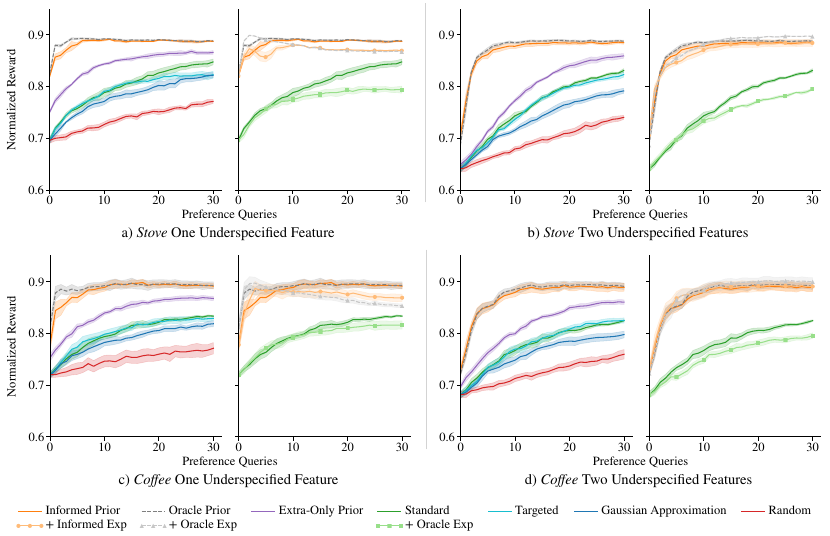}
    \caption{Average normalized reward versus number of preference queries in \textit{Stove} (a, b) and \textit{Coffee} (c, d), with one (a, c) or two (b, d) features initially underspecified. Results are averaged over all four choices of the underspecified feature (a, c) and all six pairs of underspecified features (b, d). Curves show the mean and shaded bands $\pm 1$ SE across seeds. Left panels compare methods without explanations; right panels compare explanation variants alongside their base methods.}
    \label{fig:kitchen_reward}
\end{figure*}

The informed prior accelerates reward recovery across both environments and degrees of underspecification. We evaluate settings with one and then two underspecified task-relevant features, averaging over every choice of underspecified feature or pair of underspecified features (\autoref{fig:kitchen_reward}). Results for each individual configuration are provided in \autoref{fig:kitchen_reward_full_figure} of \autoref{app:simulated-full}. In both cases, the informed prior recovers the true reward faster than the baselines. The relative ordering of the methods is also stable across the single-feature and two-feature underspecification conditions (\autoref{fig:kitchen_reward}, left panels), with the largest gains generally appearing in the low-query regime ($K \leq 10$). This is practically significant as the early queries are the regime where reducing human effort matters most. Averaged over all 20 configurations, \textit{Informed Prior} outperforms \textit{Standard} by $13.0\%$ at $K=5$ and by $11.3\%$ at $K=10$. The largest gap of $30.8\%$ occurs at $K=4$ in \textit{Stove} when \textit{human} and \textit{point} are both underspecified (\autoref{fig:kitchen_reward_full_figure}b). 
\textit{Extra-Only Prior}, which concentrates only $w^\varepsilon$ near zero, improves on \textit{Standard} but remains below \textit{Informed Prior}. This suggests that much of the gain comes from removing the misleading demonstration signal on $w^u$.
\textit{Gaussian Approximation}, which replaces $P(w \mid \mathcal{D})$ with $\mathcal{N}(\mu, \Sigma)$, performs below \textit{Standard}. Since \textit{Informed Prior} builds on this same summary, its advantage stems from the block-wise modification of $\mathcal{N}(\mu, \Sigma)$.

\paragraph{Misidentified Underspecified Features}
When one or more well-optimized features are predicted as underspecified, there is a gap at $K = 0$ between \textit{Informed Prior} and \textit{Oracle Prior} (e.g., \textit{table} in \autoref{fig:kitchen_reward_full_figure}c). This gap carries over into the averaged curves, as in \textit{Coffee} with one underspecified feature (\autoref{fig:kitchen_reward}c). If a feature $\phi_l$ is incorrectly deemed underspecified, its prior moves into the underspecified block, so its prior-anchored mean is zero despite $w^{\star}_l \neq 0$. This degrades the top-$k$ selection until preferences supply enough information to substitute for the removed signal.

\paragraph{Effect of Explanations}
\textit{Informed Prior + Informed Exp} and \textit{Oracle Prior + Oracle Exp} track their non-explanation counterparts with two underspecified features (\autoref{fig:kitchen_reward}b, \autoref{fig:kitchen_reward}d), but peak at small $K$ and then decline with one (\autoref{fig:kitchen_reward}a, \autoref{fig:kitchen_reward}c). \textit{Standard + Oracle Exp} plateaus below \textit{Standard} in all four settings. As preferences accumulate, the posterior of $w^u$ tightens and $\|w^u\|$ grows beyond the value of 0, causing $\|w^o\|$ to contract under the constraint $\|w\|_2 \leq 1$. Learning from explanation-guided queries quickly recovers information about $\phi^u$, but collects no information about balancing against the originally optimized $\phi^o$. For \textit{Standard + Oracle Exp}, explanation-guided responses additionally carry no direct information about $w^\varepsilon$, so the spurious weights in $P(w \mid \mathcal{D})$ are never directly corrected. Adaptive schemes that broaden $\phi^{\mathrm{exp}}$ as $w^u$ is resolved are left to future work.

\section{User Study}

We conducted a user study to evaluate whether generating queries under the informed prior and providing explanations to guide responses improves reward learning. We also examine whether the explanations improve participants' subjective experience along three dimensions: \textit{ease} of answering queries, \textit{understanding} of their purpose, and \textit{expression} of what the robot needs to learn.

\subsection{Experiment Design}

We conducted an IRB-approved study using the \textit{Coffee} environment. The 2$\times$2 within-subjects design crossed the prior (uninformed vs.\ informed) with the query delivery (no explanation vs.\ explanation), yielding four conditions: (1) \textit{Uninformed}, (2) \textit{Informed}, (3) \textit{Uninformed + Explanation} and (4) \textit{Informed + Explanation}. 
For each condition, participants were told the four task objectives: keep the mug upright, stay close to the table, stay away from the laptop, and stay away from the human. Participants were shown three trajectories chosen under the ground-truth reward, followed by $10$ preference queries for each of two tasks, one with \textit{laptop} initially underspecified %
and the other with \textit{human}.
The explanation conditions prefaced each task's preference queries with a natural language explanation that named the underspecified feature. All conditions used the ground-truth partition $\mathbf{o}$, leaving the prior and the explanation as the only factors that vary. Reward inference ran online: the posterior was re-sampled after every response, so queries adapted to each participant's earlier answers. We rendered queries in simulation because executing arbitrary trajectories is unsafe without a filter~\cite{mehta2024unified}, and with a filter the robot only shows acceptable behavior. Reward models calibrated on such data over-reward the behaviors humans reject~\cite{tian2026position}.
Participants answered six 7-point Likert items after each task (see \autoref{app:user-study}), completed the NASA Task Load Index (NASA-TLX)~\cite{hart1988development} after each condition, and ranked the four conditions by preference at the end. We recruited $N = 12$ participants (9 female, 2 male, 1 preferred not to say; age $M = 23.8$, $SD = 2.3$).

\paragraph{Hypotheses} We expected that: \textbf{(H1)} using the informed prior will lead to improved reward recovery; \textbf{(H2)} providing explanations to guide participant responses to queries will lead to improved reward recovery; explanations would make \textbf{(H3)} queries easier to answer, \textbf{(H4)} the purpose of queries more understandable, and \textbf{(H5)} participants feel they can better express to the robot what it needs to learn; and \textbf{(H6)} participants subjectively prefer having an explanation.

\subsection{Results}

\begin{figure*}[t]
    \centering
    \includegraphics[width=\linewidth]{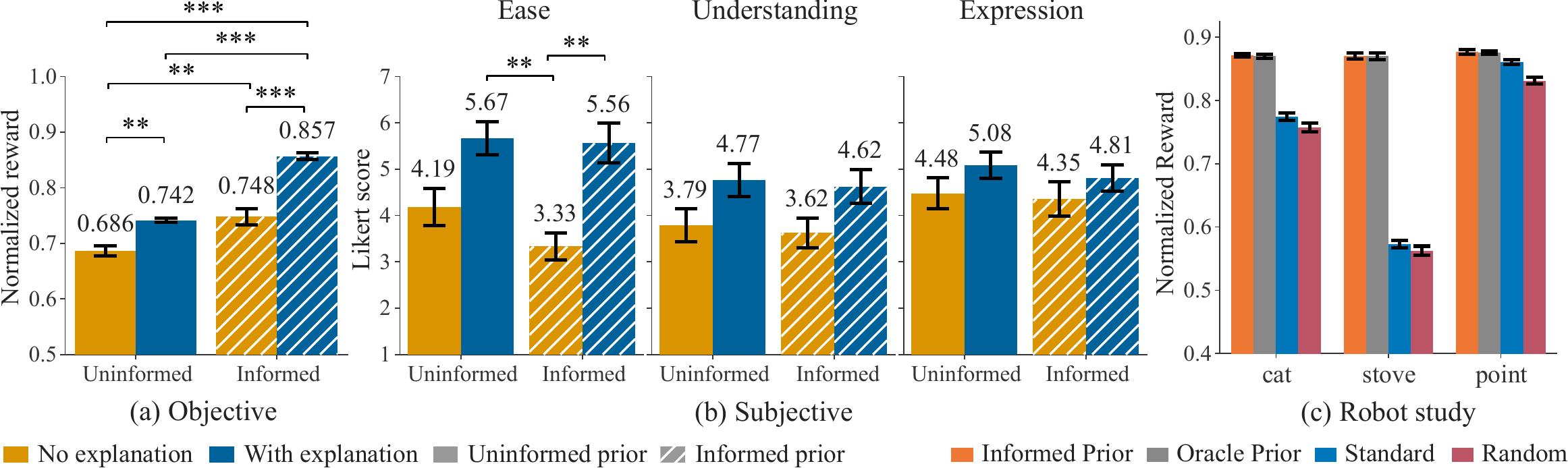}
    \caption{User study (a) normalized reward and (b) Likert score per condition. (c) Robot study mean normalized reward at $K = 10$ by feature initially underspecified. Error bars are $\pm 1$ SE.}
    \label{fig:user_study_bargraph}
\end{figure*}

\paragraph{Objective}
For each participant, condition, and task we computed normalized reward, and ran a 2$\times$2 repeated-measures ANOVA on its value at $K = 10$ (averaged across the two tasks) with prior and explanation as within-subjects factors. We found a significant main effect of prior ($F(1, 11) = 126.68$, $p < 0.001$, $\eta_p^2 = 0.920$), supporting \textbf{H1}. We also found a significant main effect of explanation ($F(1, 11) = 71.92$, $p < 0.001$, $\eta_p^2 = 0.867$), supporting \textbf{H2}. The prior $\times$ explanation interaction was also significant and positive ($F(1, 11) = 15.33$, $p = 0.002$, $\eta_p^2 = 0.582$).
\autoref{fig:user_study_bargraph}a plots the normalized reward by condition.

\paragraph{Subjective}
We ran a 2$\times$2 repeated-measures ANOVA on each Likert scale with prior and explanation as within-subjects factors. The explanation main effect was significant on \textit{ease} ($F(1, 11) = 15.31$, $p = 0.002$, $\eta_p^2 = 0.582$), supporting \textbf{H3}, and on \textit{understanding} ($F(1, 11) = 6.66$, $p = 0.026$, $\eta_p^2 = 0.377$), supporting \textbf{H4}, but not on \textit{expression} ($F(1, 11) = 2.96$, $p = 0.113$, $\eta_p^2 = 0.212$), yielding no support for \textbf{H5}. \autoref{fig:user_study_bargraph}b plots the scores by condition. We also assessed  perceived workload with a 2$\times$2 repeated-measures ANOVA with prior and explanation as within-subjects factors. Explanation significantly reduced perceived workload ($F(1, 11) = 16.37$, $p = 0.002$, $\eta_p^2 = 0.598$), while prior had no effect on workload ($F(1, 11) < 0.01$, $p = 0.985$, $\eta_p^2 < 0.001$).
A Friedman test on the rankings was not significant ($\chi^2(3) = 4.30$, $p = 0.231$, $N = 12$), giving no support for \textbf{H6} (mean ranks, where 1 is most preferred: \textit{Informed + Explanation} = 2.08, \textit{Uninformed + Explanation} = 2.17, \textit{Uninformed} = 2.75, \textit{Informed} = 3.00). 
While some participants valued directed attention (P4 noted that ``getting a cue about which feature to attend to is helpful,'' and P6 found it ``definitely easier to teach the robot when [they] knew what it was unclear about''), others felt constrained by focusing on a single dimension (P11 wasn't ``sure how much to prioritize [the feature] over the other factors,'' and P8 instead ``liked balancing the [features]'' themselves).
These responses suggest a trade-off: explanations help users focus their attention, but can conflict with a desire to balance multiple objectives, motivating attention schemes that broaden as features are resolved.

\paragraph{Demonstration-based Feedback}
Our approach assumes that preference queries resolve underspecification with less effort than additional demonstrations.
To test this, we compared NASA-TLX responses from our study's explanation conditions against existing data from~\citet{merker2026robots}, where a separate cohort of participants gave corrective kinesthetic demonstrations in response to natural-language explanations ($N = 12$ each, between subjects).
Physical demand was lower for preference queries ($M = 3.8$, $SE = 1.9$) than for kinesthetic demonstrations ($M = 31.7$, $SE = 6.9$; $t(12.6) = 3.89$, $p = 0.002$). Overall workload was also lower ($M = 17.4$ vs.\ $M = 24.2$), though not significantly ($t(22.0) = 1.31$, $p = 0.205$), and the gap shrinks once physical demand is excluded ($M = 20.1$ vs.\ $M = 22.7$; $t(21.3) = 0.45$, $p = 0.658$). Preference queries thus remove the physical cost of kinesthetic teaching, while we found no evidence of increased cognitive workload.

\section{Real Robot Examples}
\label{sec:real_robot}

We use a Franka Emika FR3 arm carrying a knife in a kitchen (\autoref{fig:environments}c). We model the kitchen in PyBullet to generate the trajectory bank and run reward learning offline. The task-relevant features are \textit{stove} ($xy$-distance to cooktop), \textit{cat} ($xy$-distance to cat), and \textit{point} (how directly the knife's extension axis points at the cat), alongside three distractors: $xy$-distances to a \textit{mug}, \textit{lego}, and \textit{rubik's cube}. The ground-truth reward keeps the knife clear of both the stove and the cat and avoids pointing the blade at the cat.
We compare \textit{Informed Prior}, \textit{Oracle Prior}, \textit{Standard}, and \textit{Random}, reporting normalized reward at $K=10$ (\autoref{fig:user_study_bargraph}c). 
Since normalized reward doesn't reveal whether gaps between methods correspond to behavior a person wouldn't like, we execute the top test trajectories on the physical arm to demonstrate that the learned rewards translate to sensible behavior (see \autoref{app:real-robot}).

\section{Conclusion}

We present an approach for resolving underspecification in reward learning that shifts the burden from the human to the robot. Our insight is that the same information that indicates which features were underspecified can also be used to ask better questions about them. The robot uses this diagnosis to shape an informed prior for active preference learning and prefaces its queries with an explanation that directs the human's attention to those features. In simulation and in a user study, the informed prior recovers the true reward faster than standard active preference learning.

\paragraph{Limitations}

The method takes the feature partition as input, so errors in the upstream diagnosis propagate into the informed prior. Updating the partition online as preferences arrive is a natural extension we leave to future work. We also summarize the demonstration-derived posterior as a Gaussian, discarding its higher-order structure. The attention scheme is similarly coarse: each query either directs the human's attention exclusively to the underspecified features or gives no explanation at all. As features become resolved, the human's attention should expand to allow later queries to also inform the well-optimized features. Softening the attention model is simple to simulate, but when and how much to broaden, and whether changing the instruction mid-task erodes ease and understanding, requires substantial human validation that we leave to future work. In addition, our experiments focused on tabletop manipulation with a small set of task-relevant features. Evaluating our approach on larger feature sets or dynamic objects would further test robustness and generality.

\clearpage
\acknowledgments{
This work has taken place in the Collaborative Learning and Autonomy Research Lab (CLEAR Lab) at MIT. CLEAR research is supported in part by the National Science Foundation (OIA-2535195) and the MIT Schwarzman College of Computing’s Social and Ethical Responsibilities of Computing program. Any opinions, findings, and conclusions expressed in this material are those of the authors and do not necessarily reflect the views of the sponsors.
}

\bibliography{example}  %

\begin{thebibliography}{33}
\providecommand{\natexlab}[1]{#1}
\providecommand{\url}[1]{\texttt{#1}}
\expandafter\ifx\csname urlstyle\endcsname\relax
  \providecommand{\doi}[1]{doi: #1}\else
  \providecommand{\doi}{doi: \begingroup \urlstyle{rm}\Url}\fi

\bibitem[Sweller(1988)]{sweller1988cognitive}
J.~Sweller.
\newblock Cognitive load during problem solving: Effects on learning.
\newblock \emph{Cognitive Science}, 12\penalty0 (2):\penalty0 257--285, 1988.
\newblock \doi{10.1207/s15516709cog1202\_4}.

\bibitem[Baddeley(1992)]{baddeley1992working}
A.~Baddeley.
\newblock Working memory.
\newblock \emph{Science}, 255\penalty0 (5044):\penalty0 556--559, 1992.
\newblock \doi{10.1126/science.1736359}.

\bibitem[Merker et~al.(2026)Merker, Walker, and Bobu]{merker2026robots}
H.~Merker, N.~Walker, and A.~Bobu.
\newblock Robots that know what to ask: Recovering misaligned rewards through
  targeted explanations.
\newblock In \emph{Proceedings of Robotics: Science and Systems}, Sydney,
  Australia, 2026.
\newblock \doi{10.15607/RSS.2026.XXII.116}.

\bibitem[Peng et~al.(2023)Peng, Netanyahu, Ho, Shu, Bobu, Shah, and
  Agrawal]{peng2023DFA}
A.~Peng, A.~Netanyahu, M.~K. Ho, T.~Shu, A.~Bobu, J.~Shah, and P.~Agrawal.
\newblock Diagnosis, feedback, adaptation: {A} human-in-the-loop framework for
  test-time policy adaptation.
\newblock In \emph{International Conference on Machine Learning, {ICML} 2023,
  23-29 July 2023, Honolulu, Hawaii, {USA}}, volume 202 of \emph{Proceedings of
  Machine Learning Research}, pages 27630--27641. {PMLR}, 2023.
\newblock URL \url{https://proceedings.mlr.press/v202/peng23c.html}.

\bibitem[Sadigh et~al.(2017)Sadigh, Dragan, Sastry, and
  Seshia]{sadigh2017active}
D.~Sadigh, A.~D. Dragan, S.~Sastry, and S.~A. Seshia.
\newblock Active preference-based learning of reward functions.
\newblock In \emph{Robotics: Science and Systems XIII, Massachusetts Institute
  of Technology, Cambridge, Massachusetts, USA, July 12-16, 2017}, 2017.
\newblock \doi{10.15607/RSS.2017.XIII.053}.

\bibitem[Biyik et~al.(2019)Biyik, Palan, Landolfi, Losey, and
  Sadigh]{biyik2019asking}
E.~Biyik, M.~Palan, N.~C. Landolfi, D.~P. Losey, and D.~Sadigh.
\newblock Asking easy questions: {A} user-friendly approach to active reward
  learning.
\newblock In \emph{3rd Annual Conference on Robot Learning, CoRL 2019, Osaka,
  Japan, October 30 - November 1, 2019, Proceedings}, Proceedings of Machine
  Learning Research, pages 1177--1190. {PMLR}, 2019.
\newblock URL \url{http://proceedings.mlr.press/v100/b-iy-ik20a.html}.

\bibitem[Peng et~al.(2024)Peng, Sun, Shu, and
  Abel]{peng2024pragmaticfeaturepreferences}
A.~Peng, Y.~Sun, T.~Shu, and D.~Abel.
\newblock Pragmatic feature preferences: Learning reward-relevant preferences
  from human input.
\newblock In \emph{Forty-first International Conference on Machine Learning,
  {ICML} 2024, Vienna, Austria, July 21-27, 2024}, Proceedings of Machine
  Learning Research, pages 40258--40270. {PMLR}, 2024.
\newblock URL \url{https://proceedings.mlr.press/v235/peng24d.html}.

\bibitem[Biyik et~al.(2022)Biyik, Losey, Palan, Landolfi, Shevchuk, and
  Sadigh]{biyik2022learningreward}
E.~Biyik, D.~P. Losey, M.~Palan, N.~C. Landolfi, G.~Shevchuk, and D.~Sadigh.
\newblock Learning reward functions from diverse sources of human feedback:
  Optimally integrating demonstrations and preferences.
\newblock \emph{Int. J. Robotics Res.}, 41\penalty0 (1):\penalty0 45--67, 2022.
\newblock \doi{10.1177/02783649211041652}.

\bibitem[Habibian et~al.(2022)Habibian, Jonnavittula, and
  Losey]{habibian2022heres}
S.~Habibian, A.~Jonnavittula, and D.~P. Losey.
\newblock \emph{Here's What I've Learned: } asking questions that reveal reward
  learning.
\newblock \emph{{ACM} Trans. Hum. Robot Interact.}, 11\penalty0 (4):\penalty0
  40:1--40:28, 2022.
\newblock \doi{10.1145/3526107}.

\bibitem[Abbeel and Ng(2004)]{abbeel2004apprenticeship}
P.~Abbeel and A.~Y. Ng.
\newblock Apprenticeship learning via inverse reinforcement learning.
\newblock In \emph{Machine Learning, Proceedings of the Twenty-first
  International Conference {(ICML} 2004), Banff, Alberta, Canada, July 4-8,
  2004}, volume~69 of \emph{{ACM} International Conference Proceeding Series}.
  {ACM}, 2004.
\newblock \doi{10.1145/1015330.1015430}.

\bibitem[Ziebart et~al.(2008)Ziebart, Maas, Bagnell, Dey,
  et~al.]{ziebart2008maximum}
B.~D. Ziebart, A.~L. Maas, J.~A. Bagnell, A.~K. Dey, et~al.
\newblock Maximum entropy inverse reinforcement learning.
\newblock In \emph{AAAI}, volume~8, pages 1433--1438. Chicago, IL, USA, 2008.

\bibitem[Laidlaw and Dragan(2022)]{laidlaw2022boltzmann}
C.~Laidlaw and A.~D. Dragan.
\newblock The boltzmann policy distribution: Accounting for systematic
  suboptimality in human models.
\newblock In \emph{The Tenth International Conference on Learning
  Representations, {ICLR} 2022, Virtual Event, April 25-29, 2022}.
  OpenReview.net, 2022.
\newblock URL \url{https://openreview.net/forum?id=\_l\_QjPGN5ye}.

\bibitem[Shah et~al.(2019)Shah, Gundotra, Abbeel, and
  Dragan]{shah2019feasibility}
R.~Shah, N.~Gundotra, P.~Abbeel, and A.~D. Dragan.
\newblock On the feasibility of learning, rather than assuming, human biases
  for reward inference.
\newblock In \emph{Proceedings of the 36th International Conference on Machine
  Learning, {ICML} 2019, 9-15 June 2019, Long Beach, California, {USA}},
  Proceedings of Machine Learning Research, pages 5670--5679. {PMLR}, 2019.
\newblock URL \url{http://proceedings.mlr.press/v97/shah19a.html}.

\bibitem[Bobu et~al.(2018)Bobu, Bajcsy, Fisac, and
  Dragan]{bobu2018misspecified}
A.~Bobu, A.~Bajcsy, J.~F. Fisac, and A.~D. Dragan.
\newblock Learning under misspecified objective spaces.
\newblock In \emph{2nd Annual Conference on Robot Learning, CoRL 2018,
  Z{\"{u}}rich, Switzerland, 29-31 October 2018, Proceedings}, Proceedings of
  Machine Learning Research, pages 796--805. {PMLR}, 2018.
\newblock URL \url{http://proceedings.mlr.press/v87/bobu18a.html}.

\bibitem[Bobu et~al.(2020)Bobu, Bajcsy, Fisac, Deglurkar, and
  Dragan]{bobu2020quantifying}
A.~Bobu, A.~Bajcsy, J.~F. Fisac, S.~Deglurkar, and A.~D. Dragan.
\newblock Quantifying hypothesis space misspecification in learning from
  human-robot demonstrations and physical corrections.
\newblock \emph{{IEEE} Trans. Robotics}, 36\penalty0 (3):\penalty0 835--854,
  2020.
\newblock \doi{10.1109/TRO.2020.2971415}.

\bibitem[Zurek et~al.(2021)Zurek, Bobu, Brown, and Dragan]{zurek2021confidence}
M.~Zurek, A.~Bobu, D.~S. Brown, and A.~D. Dragan.
\newblock Situational confidence assistance for lifelong shared autonomy.
\newblock In \emph{International Conference on Robotics and Automation (ICRA)},
  2021.

\bibitem[Louren{\c{c}}o et~al.(2023)Louren{\c{c}}o, Bobu, Rojas, and
  Wahlberg]{lourenco2023diagnosing}
I.~Louren{\c{c}}o, A.~Bobu, C.~R. Rojas, and B.~Wahlberg.
\newblock Diagnosing and repairing feature representations under distribution
  shifts.
\newblock In \emph{62nd {IEEE} Conference on Decision and Control, {CDC} 2023,
  Singapore, December 13-15, 2023}, pages 3638--3645. {IEEE}, 2023.
\newblock \doi{10.1109/CDC49753.2023.10383644}.
\newblock URL \url{https://doi.org/10.1109/CDC49753.2023.10383644}.

\bibitem[Mahmud et~al.(2025)Mahmud, Nakamura, and Zilberstein]{mahmud2025maple}
S.~Mahmud, M.~Nakamura, and S.~Zilberstein.
\newblock {MAPLE:} {A} framework for active preference learning guided by large
  language models.
\newblock In \emph{Thirty-Ninth {AAAI} Conference on Artificial Intelligence,
  Thirty-Seventh Conference on Innovative Applications of Artificial
  Intelligence, Fifteenth Symposium on Educational Advances in Artificial
  Intelligence, {AAAI} 2025, Philadelphia, PA, USA, February 25 - March 4,
  2025}, pages 27518--27528. {AAAI} Press, 2025.
\newblock \doi{10.1609/AAAI.V39I26.34964}.

\bibitem[Bobu et~al.(2022)Bobu, Wiggert, Tomlin, and Dragan]{bobu2022inducing}
A.~Bobu, M.~Wiggert, C.~J. Tomlin, and A.~D. Dragan.
\newblock Inducing structure in reward learning by learning features.
\newblock \emph{The International Journal of Robotics Research}, 41\penalty0
  (5):\penalty0 497--518, 2022.
\newblock \doi{10.1177/02783649221078031}.

\bibitem[Peng et~al.(2024)Peng, Li, Sucholutsky, Kumar, Shah, Andreas, and
  Bobu]{peng24algae}
A.~Peng, B.~Z. Li, I.~Sucholutsky, N.~Kumar, J.~Shah, J.~Andreas, and A.~Bobu.
\newblock Adaptive language-guided abstraction from contrastive explanations.
\newblock In \emph{Conference on Robot Learning, 6-9 November 2024, Munich,
  Germany}, volume 270 of \emph{Proceedings of Machine Learning Research},
  pages 3425--3438. {PMLR}, 2024.
\newblock URL \url{https://proceedings.mlr.press/v270/peng25c.html}.

\bibitem[Bajcsy et~al.(2017)Bajcsy, Losey, O'Malley, and
  Dragan]{bajcsy2017learning}
A.~Bajcsy, D.~P. Losey, M.~K. O'Malley, and A.~D. Dragan.
\newblock Learning robot objectives from physical human interaction.
\newblock In \emph{1st Annual Conference on Robot Learning, CoRL 2017, Mountain
  View, California, USA, November 13-15, 2017, Proceedings}, volume~78 of
  \emph{Proceedings of Machine Learning Research}, pages 217--226. {PMLR},
  2017.
\newblock URL \url{http://proceedings.mlr.press/v78/bajcsy17a.html}.

\bibitem[Sumers et~al.(2021)Sumers, Ho, Hawkins, Narasimhan, and
  Griffiths]{sumers2021learning}
T.~R. Sumers, M.~K. Ho, R.~X.~D. Hawkins, K.~Narasimhan, and T.~L. Griffiths.
\newblock Learning rewards from linguistic feedback.
\newblock In \emph{Thirty-Fifth {AAAI} Conference on Artificial Intelligence,
  {AAAI} 2021, Thirty-Third Conference on Innovative Applications of Artificial
  Intelligence, {IAAI} 2021, The Eleventh Symposium on Educational Advances in
  Artificial Intelligence, {EAAI} 2021, Virtual Event, February 2-9, 2021},
  pages 6002--6010. {AAAI} Press, 2021.
\newblock \doi{10.1609/AAAI.V35I7.16749}.

\bibitem[Jaynes(1957)]{jaynes1957infotheory}
E.~T. Jaynes.
\newblock Information theory and statistical mechanics.
\newblock \emph{Phys. Rev.}, 106:\penalty0 620--630, May 1957.
\newblock \doi{10.1103/PhysRev.106.620}.

\bibitem[Von~Neumann and Morgenstern(1945)]{von1945theory}
J.~Von~Neumann and O.~Morgenstern.
\newblock \emph{Theory of games and economic behavior}.
\newblock Princeton University Press Princeton, NJ, 1945.

\bibitem[Baker et~al.(2007)Baker, Tenenbaum, and Saxe]{baker2007goal}
C.~L. Baker, J.~B. Tenenbaum, and R.~R. Saxe.
\newblock Goal inference as inverse planning.
\newblock In \emph{Proceedings of the Annual Meeting of the Cognitive Science
  Society}, volume~29, 2007.

\bibitem[Minka(2001)]{minka2001expectation}
T.~P. Minka.
\newblock Expectation propagation for approximate bayesian inference.
\newblock In \emph{{UAI} '01: Proceedings of the 17th Conference in Uncertainty
  in Artificial Intelligence, University of Washington, Seattle, Washington,
  USA, August 2-5, 2001}, pages 362--369. Morgan Kaufmann, 2001.

\bibitem[Bradley and Terry(1952)]{bradley1952rank}
R.~A. Bradley and M.~E. Terry.
\newblock Rank analysis of incomplete block designs: I. the method of paired
  comparisons.
\newblock \emph{Biometrika}, 39\penalty0 (3/4):\penalty0 324--345, 1952.
\newblock ISSN 00063444, 14643510.
\newblock URL \url{http://www.jstor.org/stable/2334029}.

\bibitem[Cover and Thomas(2012)]{cover2012elements}
T.~M. Cover and J.~A. Thomas.
\newblock \emph{Elements of Information Theory}.
\newblock John Wiley \& Sons, 2012.

\bibitem[Mehta and Losey(2024)]{mehta2024unified}
S.~A. Mehta and D.~P. Losey.
\newblock Unified learning from demonstrations, corrections, and preferences
  during physical human-robot interaction.
\newblock \emph{{ACM} Trans. Hum. Robot Interact.}, 13\penalty0 (3):\penalty0
  39:1--39:25, 2024.
\newblock \doi{10.1145/3623384}.

\bibitem[Tian et~al.(2026)Tian, Wu, and Bajcsy]{tian2026position}
R.~Tian, Y.~Wu, and A.~Bajcsy.
\newblock Position: Good embodied reward models need bad behavior data.
\newblock In \emph{Proceedings of the 43rd International Conference on Machine
  Learning}, Seoul, South Korea, 2026.
\newblock arXiv:2606.01036.

\bibitem[Hart and Staveland(1988)]{hart1988development}
S.~G. Hart and L.~E. Staveland.
\newblock Development of {NASA-TLX} (task load index): Results of empirical and
  theoretical research.
\newblock \emph{Advances in Psychology}, 52:\penalty0 139--183, 1988.
\newblock \doi{10.1016/S0166-4115(08)62386-9}.

\bibitem[Singmann et~al.(2025)Singmann, Bolker, Westfall, Aust, and
  Ben-Shachar]{singmann2025afex}
H.~Singmann, B.~Bolker, J.~Westfall, F.~Aust, and M.~S. Ben-Shachar.
\newblock \emph{afex: Analysis of Factorial Experiments}, 2025.
\newblock URL \url{https://CRAN.R-project.org/package=afex}.
\newblock R package version 1.5-1.

\bibitem[Lenth and Piaskowski(2026)]{lenth2026emmeans}
R.~V. Lenth and J.~Piaskowski.
\newblock \emph{emmeans: Estimated Marginal Means, aka Least-Squares Means},
  2026.
\newblock URL \url{https://CRAN.R-project.org/package=emmeans}.
\newblock R package version 2.0.3.

\end{thebibliography}

\clearpage

\appendix

\section{Experimental Details}\label{app:experimental-details}

\paragraph{Identifying Irrelevant Features}
The candidate feature set $\phi$ is not guaranteed to be task-relevant in its entirety. Features that have no bearing on the task carry no signal in $\mathcal{D}$, so the demonstration-derived posterior along these dimensions is shaped not by the human's intent but by spurious correlations in the data. We use an LLM to identify these features. For each candidate feature $\phi_l$, we issue $N$ independent queries to the LLM. Each query supplies the task description and the full candidate feature set, and asks the model to label $\phi_l$ as either relevant or irrelevant. If a majority of the $N$ responses vote irrelevant, we assign $\phi_l$ to $\phi^\varepsilon$ and record the LLM's confidence as $c_l \in [0.5, 1]$, the fraction of the $N$ responses that voted irrelevant.

\autoref{sec:prior} defines the prior's precision on $\phi^\varepsilon$ as $\tilde\Lambda^{\varepsilon\varepsilon} = \mathrm{diag}\bigl((\lambda_l)_{l \in \phi^\varepsilon}\bigr)$, so each $\lambda_l \geq 0$ controls how tightly the prior pulls $w_l$ toward zero. $\lambda_l$ should match the concentration of $P(w \mid \mathcal{D})$ on $w^o$ when the LLM is fully confident that $\phi_l$ is irrelevant, and weaken when the LLM is less confident. We set $\lambda_l = c_l \, \tilde\tau$, where $\tilde\tau = \mathrm{median}\bigl\{1/\Sigma_{ll} : l \in \phi^o\bigr\}$ is the median precision among the well-optimized features and $\Sigma_{ll}$ is the diagonal entry of the Gaussian summary $\mathcal{N}(\mu, \Sigma) \approx P(w \mid \mathcal{D})$. The scalar $\tilde\tau$ sets a default concentration strength comparable to the prior's concentration on $w^o$, and the factor $c_l$ scales this strength down when the LLM is less confident that $\phi_l$ is irrelevant. We take $\tilde\Lambda^{\varepsilon\varepsilon}$ to be diagonal because the LLM evaluates each feature independently, providing no signal about cross-feature correlations within $\phi^\varepsilon$.

\paragraph{LLM Prompts}
\label{app:llm-prompts}

We use OpenAI's \texttt{gpt-5.2} with a structured JSON output schema and reasoning effort set to high. The prompts for the \textit{Stove}, \textit{Coffee}, and \textit{Kitchen} environments are shown below. For each candidate feature $\phi_l$, we issue the corresponding environment's prompt $N = 10$ times, with \{feature\} and \{feature\_description\} replaced by the name and short description of $\phi_l$.

\begin{promptbox}[title=\textit{Stove} prompt]
\setlength{\parindent}{0pt}
\setlength{\parskip}{0.5\baselineskip}

The following is an instruction for a task that the agent needs to complete. instruction = A 7-DoF Jaco robot arm is carrying a utensil above a tabletop environment. The arm is carrying a utensil with a sharp, pointed tip. Various objects are on the table, including a stove, a sponge, a banana, a tomato soup can, and a sugar box. A human is right next to the table. A ceiling rack is right above the table.

The current available feature names are: human (distance from end-effector to human) stove (distance from end-effector to stove) point (how much the utensil held by the end-effector points toward the human) rack ($z$-distance of the end-effector below the ceiling rack) sponge (distance from end-effector to sponge) banana (distance from end-effector to banana) soup (distance from end-effector to tomato soup can) sugar (distance from end-effector to sugar box)

We are evaluating whether \{feature\} (\{feature\_description\}) is part of the set of features that a typical human would consider necessary for this task.

Please return either a 1 or 0 for the following feature: feature = \{feature\} (\{feature\_description\}).

A feature should be labeled 1 only if a typical human would expect that ignoring it would commonly lead to a clearly undesirable outcome in this specific task. If it is unclear whether the feature would meaningfully affect behavior in this context, then it should be labeled 0.

Please reason step by step internally.

Provide your answer as ONLY JSON in one of these two forms: \{"label": 0\} or \{"label": 1\}. Do not include any surrounding explanation or text.
\end{promptbox}

\begin{promptbox}[title=\textit{Coffee} prompt]
\setlength{\parindent}{0pt}
\setlength{\parskip}{0.5\baselineskip}

The following is an instruction for a task that the agent needs to complete. instruction = A 7-DoF Jaco robot arm is carrying a coffee cup above a tabletop environment. The arm is carrying a fragile ceramic coffee cup that is at risk of slipping. Various objects are on the table. A human is right next to the table.

The current available feature names are: laptop (distance from end-effector to laptop) table (height of end-effector above table surface) human (distance from end-effector to human) coffee (coffee cup orientation (tilt from upright)) apple (distance from end-effector to apple) bowl (distance from end-effector to bowl) fork (distance from end-effector to fork) marker (distance from end-effector to marker)

We are evaluating whether \{feature\} (\{feature\_description\}) is part of the set of features that a typical human would consider necessary for this task.

Please return either a 1 or 0 for the following feature: feature = \{feature\} (\{feature\_description\}).

A feature should be labeled 1 only if a typical human would expect that ignoring it would commonly lead to a clearly undesirable outcome in this specific task. If it is unclear whether the feature would meaningfully affect behavior in this context, then it should be labeled 0.

Please reason step by step internally.

Provide your answer as ONLY JSON in one of these two forms: \{"label": 0\} or \{"label": 1\}. Do not include any surrounding explanation or text.
\end{promptbox}

\begin{promptbox}[title=\textit{Kitchen} prompt]
\setlength{\parindent}{0pt}
\setlength{\parskip}{0.5\baselineskip}

The following is an instruction for a task that the agent needs to complete. instruction = A 7-DoF Franka robot arm is carrying a utensil in a kitchen environment. The end-effector is carrying a knife with a sharp, pointed tip. On the counter there is a stove with burners and a live cat sitting nearby. Above the counter there are cabinets and a microwave. Also on the counter is a red mug, a lego block, and a rubik's cube.

The current available feature names are: cat (distance from end-effector to a live cat) stove (distance from end-effector to stove) point (how much the knife held by the end-effector points toward the live cat on the counter) mug (distance from end-effector to a mug) lego (distance from end-effector to a lego block) rubiks (distance from end-effector to a rubik's cube)

We are evaluating whether \{feature\} (\{feature\_description\}) is part of the set of features that a typical human would consider necessary for this task.

Please return either a 1 or 0 for the following feature: feature = \{feature\} (\{feature\_description\}).

A feature should be labeled 1 only if a typical human would expect that ignoring it would commonly lead to a clearly undesirable outcome in this specific task. If it is unclear whether the feature would meaningfully affect behavior in this context, then it should be labeled 0.

Please reason step by step internally.

Provide your answer as ONLY JSON in one of these two forms: \{"label": 0\} or \{"label": 1\}. Do not include any surrounding explanation or text.
\end{promptbox}

\section{Per-Configuration Simulation Results}\label{app:simulated-full}

\begin{figure*}[t]
    \centering
    \includegraphics[width=\linewidth]{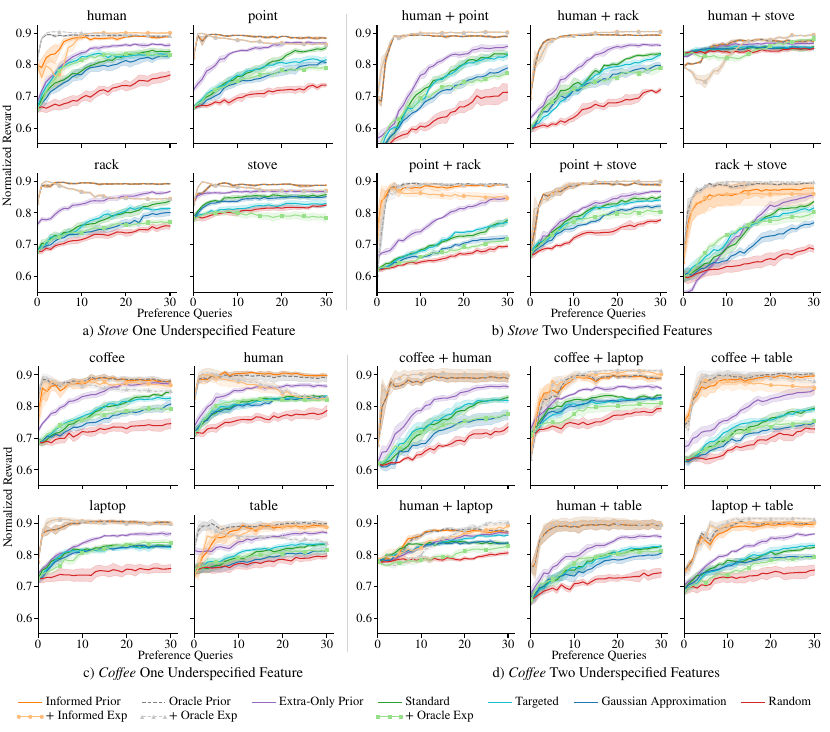}
    \caption{Normalized reward with (a, c) one underspecified feature and (b, d) two underspecified features for the \textit{Stove} (a, b) and \textit{Coffee} (c, d) environments. Across conditions, the informed prior recovers the true reward faster than the baselines, with the largest gains in the low-query regime.}
    \label{fig:kitchen_reward_full_figure}
\end{figure*}

We report each of the 20 configurations individually in \autoref{fig:kitchen_reward_full_figure}, complementing the averaged results in \autoref{fig:kitchen_reward}. Each environment contributes four configurations with a single underspecified feature (a, c) and six with a pair of underspecified features (b, d). Curves show the mean and the shaded regions the standard error across 5 seeds.

\section{User Study Details}\label{app:user-study}

\paragraph{Procedure} 
After providing informed consent, participants were familiarized with the task and the preference-query interface. They were told that the robot was learning to carry a mug of coffee across a cluttered table and that it had not yet learned to do so correctly. Four task objectives were explicitly stated: (1) keeping the mug upright, (2) staying close to the table, (3) staying away from the laptop, and (4) staying away from the person. These objectives were repeated before each condition. Participants did not provide any demonstrations: the underspecified teaching signal that began each task was generated by a simulated Boltzmann-rational demonstrator (rationality $\beta^u=0$ on the task's underspecified feature) and held fixed across all participants and conditions using a single random seed, so that every participant began each task from an identical reward belief and contributed only preference responses.

\begin{figure}
    \centering
    \includegraphics[width=\linewidth]{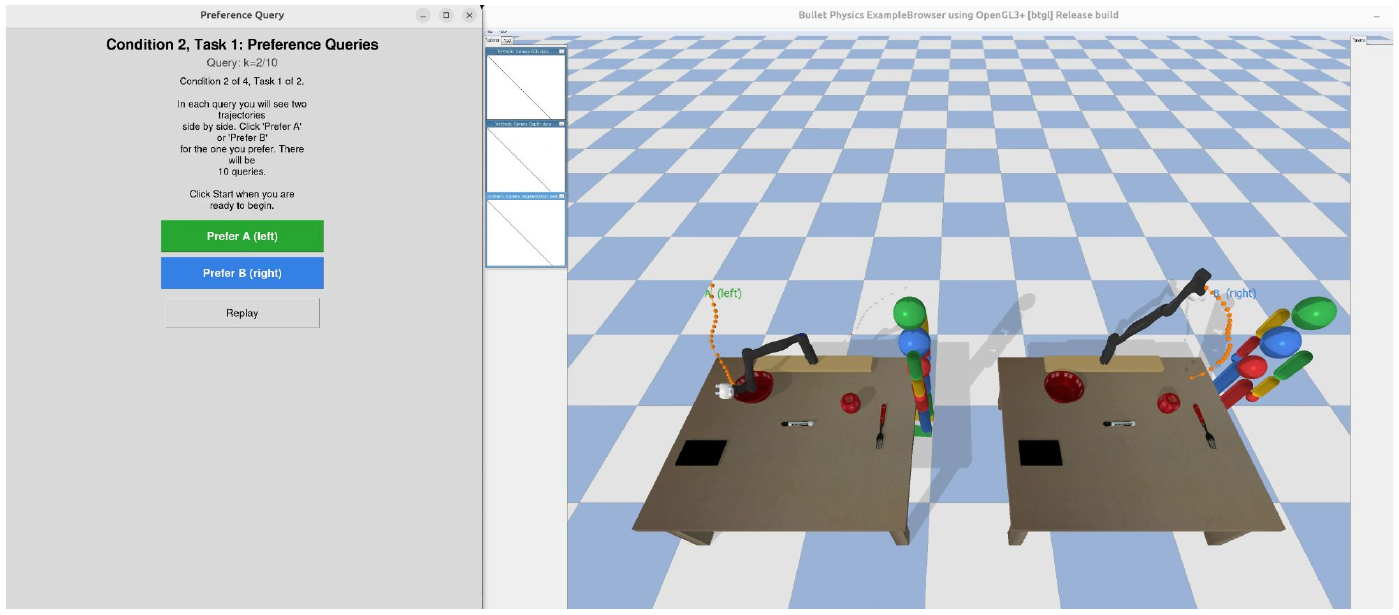}
    \caption{An example preference query from the user study in the \textit{Coffee} environment: participants are shown two trajectories side by side and select the one they prefer.}
    \label{fig:user_study_screenshot}
\end{figure}

The study crossed the prior (uninformed vs. informed) with the query delivery (no explanation vs. explanation), yielding four conditions: \textit{Uninformed}, \textit{Informed}, \textit{Uninformed + Explanation}, and \textit{Informed + Explanation}. Each condition began with a short familiarization phase in which the participant watched three example trajectories drawn under the ground-truth reward. The participant then completed 2 tasks, one with \textit{laptop} initially underspecified and one with \textit{human} initially underspecified, each consisting of 10 preference queries. The two framings were:
\begin{itemize}
\item \textit{No explanation} (\textit{Uninformed}, \textit{Informed}): ``In each query you will see two trajectories side by side. Click `Prefer A' or `Prefer B' for the one you prefer.''
\item \textit{Explanation} (\textit{Uninformed + Explanation}, \textit{Informed + Explanation}): the same instructions, followed by ``The robot is uncertain about how to handle \texttt{[feature]}. Please focus on this feature when comparing the two trajectories,'' where \texttt{[feature]} was ``distance to the laptop'' or ``distance to the human'' depending on the task.
\end{itemize}
Preference queries were presented through a PyBullet interface that played the two candidate trajectories side by side (\autoref{fig:user_study_screenshot}). Participants chose between them with ``Prefer A (left)'' and ``Prefer B (right)'' buttons, and could re-watch both trajectories as often as they wished with a ``Replay'' button. Condition order was counterbalanced with a Williams Latin square of size four. The order of the two tasks within each condition was independently counterbalanced so that, within every condition, the participants split evenly between laptop-first and human-first. After each task, participants answered the six 7-point Likert items in~\autoref{tab:survey_scales}; after each condition, they completed the NASA Task Load Index (NASA-TLX)~\cite{hart1988development}; and at the end of the session, they ranked the four conditions from most to least preferred.

\begin{table}[htbp]
\centering
\caption{ANOVA results for the main effect of explanation on Likert scales.}
\label{tab:survey_scales}
\setlength{\tabcolsep}{4pt}
\renewcommand{\arraystretch}{1.15}
\small
\begin{tabularx}{\linewidth}{l p{8.5cm} c c c}
\hline
\textbf{Scale} & \textbf{Questions} & \textbf{Cronbach's $\alpha$} & \textbf{\textit{F}} & \textbf{\textit{p}} \\
\hline
Ease & I found it easy to choose between the two trajectories in each query. & \multirow{2}{*}{0.94} & \multirow{2}{*}{15.31} & \multirow{2}{*}{0.002} \\
              & The queries were difficult to answer. & & & \\
\hline
Under. & I understood what each query was trying to learn. & \multirow{2}{*}{0.83} & \multirow{2}{*}{6.66} & \multirow{2}{*}{0.026} \\
                & The queries seemed random or unmotivated. & & & \\
\hline
Expr. & The queries covered important parts of the task. & \multirow{2}{*}{0.75} & \multirow{2}{*}{2.96} & \multirow{2}{*}{0.113} \\
               & Queries were wasted on things that didn't matter. & & & \\
\hline
\end{tabularx}
\end{table}

\paragraph{Analysis} 
We ran every test in R. We analyzed the objective measure with a $2\times2$ repeated-measures ANOVA (\texttt{afex}~\cite{singmann2025afex}) with prior and explanation as within-subjects factors and partial eta-squared ($\eta_p^2$) as the effect-size measure. We supplemented the ANOVA with Tukey-corrected pairwise contrasts among the four conditions, computed on the estimated marginal means (\texttt{emmeans}~\cite{lenth2026emmeans}) with within-subjects Cohen's $d_z$. These contrasts showed that \textit{Informed + Explanation} significantly outperformed \textit{Uninformed} ($p < 0.001$, $d_z = 5.84$), and that \textit{Informed} ($p = 0.003$, $d_z = 1.36$) and \textit{Uninformed + Explanation} ($p = 0.001$, $d_z = 1.55$) each significantly outperformed \textit{Uninformed}, indicating that the prior and the explanation each contribute measurable gains and that combining them produces the largest gain. \textit{Informed} and \textit{Uninformed + Explanation} did not differ from each other ($p = 0.976$), suggesting that the prior alone and the explanation alone provide comparable benefit over baseline.

For the subjective measures, each negatively-worded Likert item was reverse-coded and averaged with its positively-worded partner to form the three two-item scales (\textit{ease}, \textit{understanding}, and \textit{expression}). Each scale was analyzed with its own $2\times2$ repeated-measures ANOVA. Beyond the explanation main effects in \autoref{tab:survey_scales}, the prior main effect was not significant for any scale (\textit{ease}: $F(1, 11) = 2.20$, $p = 0.166$; \textit{understanding}: $F(1, 11) = 0.83$, $p = 0.382$; \textit{expression}: $F(1, 11) = 1.07$, $p = 0.323$), and the prior $\times$ explanation interaction was significant only for \textit{ease} ($F(1, 11) = 6.67$, $p = 0.025$, $\eta_p^2 = 0.378$; \textit{understanding}: $F(1, 11) < 0.01$, $p = 0.961$; \textit{expression}: $F(1, 11) = 0.17$, $p = 0.690$).

Perceived workload was measured once per condition and summarized as the unweighted mean of the six subscales on a $0$--$100$ scale. We used a $2\times2$ repeated-measures ANOVA and supplemented it with Tukey-corrected pairwise contrasts on the estimated marginal means. These contrasts localized the workload reduction to the two explanation conditions. \textit{Uninformed + Explanation} ($M = 17.6$, $p = 0.044$, $d_z = 0.89$) and \textit{Informed + Explanation} ($M = 17.2$, $p = 0.045$, $d_z = 0.89$) each significantly lowered workload relative to \textit{Uninformed} ($M = 24.7$), whereas \textit{Informed} did not ($M = 25.1$, $p = 0.999$, $d_z = -0.05$).

\section{Real Robot Execution}\label{app:real-robot}

\begin{wrapfigure}{r}{0.3\linewidth}
    \centering
    \includegraphics[width=\linewidth]{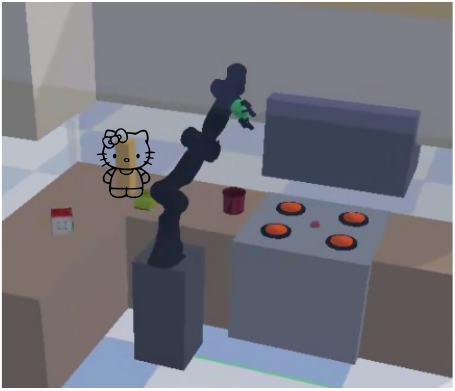}
    \caption{The simulated model of the kitchen used to generate the trajectory bank for the real robot demos. A box is used as a visual placeholder for the cat in the actual environment.}
    \label{fig:simulated_kitchen}
\end{wrapfigure}
\begin{figure}
    \centering
    \includegraphics[width=\linewidth]{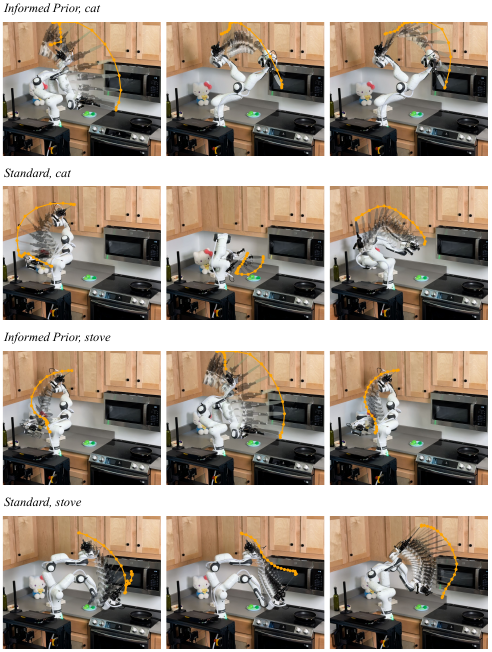}
    \caption{The three highest-ranked test trajectories executed on the Franka Emika FR3 arm after $K=10$ preference queries, for \textit{Informed Prior} and \textit{Standard} when \textit{cat} (top two rows) or \textit{stove} (bottom two rows) was initially underspecified. Each panel traces the path of the knife in orange.}
    \label{fig:real_robot_trajectories}
\end{figure}

We validate that the learned rewards translate into sensible physical behavior by executing the top trajectories on the Franka Emika FR3 arm  (\autoref{fig:environments}c). The trajectory bank is generated entirely within a PyBullet model of the kitchen (\autoref{fig:simulated_kitchen}) and all reward learning is performed offline; only the final selected trajectories are run on hardware. For each of the four methods (\textit{Informed Prior}, \textit{Oracle Prior}, \textit{Standard}, and \textit{Random}) and each of the three underspecification conditions (\textit{cat}, \textit{stove}, or \textit{point} underspecified), we take the top-ranked test trajectories after $K=10$ queries and execute them on the physical arm. Both \textit{Informed Prior} and \textit{Oracle Prior} are visually consistent across all three underspecification conditions. Whether \textit{cat}, \textit{stove}, or \textit{point} was initially underspecified, the top trajectories steer the end-effector away from both the cat and the stove while keeping the knife pointed away from the cat. The \textit{Standard} baseline mirrors \textit{Informed Prior} when \textit{point} was underspecified. However, when \textit{cat} was underspecified, the top trajectories carry the knife substantially closer to the cat; when \textit{stove} was underspecified, they likewise route the end-effector closer to the stove (\autoref{fig:real_robot_trajectories}). In neither case did the preference queries recover the clearance that the demonstrations had missed. The \textit{Random} baseline follows the same trend: visually similar to \textit{Informed Prior} when \textit{point} was underspecified, yet routing the knife closer to whichever of \textit{cat} or \textit{stove} was initially underspecified.

The similar performance of all four methods when \textit{point} is underspecified is partly a consequence of the trajectory bank, which contains a wide range of orientations. Since \textit{point} is only weakly correlated with the distance features and has the widest spread of any feature, trajectories vary widely along \textit{point}, and comparisons that differ along it are abundant. Trajectories that stay clear of both the \textit{cat} and the \textit{stove} at once are less common in the bank, so comparisons that would teach the robot to balance the two are rarer. This is a property of our kitchen, whose reachable workspace is small in $xy$; a larger scene would leave more room to stay clear of both objects at once. In addition, although the \textit{stove}-underspecified condition shows a large gap in normalized reward between methods (\autoref{fig:user_study_bargraph}c), the visible difference between some of the trajectories (\autoref{fig:real_robot_trajectories}) is less pronounced than this gap suggests. The \textit{stove} feature scores the end-effector's distance to a single reference point (cooktop center). Because the stove itself is a large object, distance to that point is only a loose proxy for how close the gripper actually appears to the stove. The end-effector can pass near the stove's physical edge while still registering a sizable distance to its center.

\end{document}